%% file: main.tex
\documentclass{article}

\PassOptionsToPackage{numbers, compress}{natbib}
\usepackage[preprint]{neurips_2026}      

\usepackage[utf8]{inputenc}
\usepackage[T1]{fontenc}
\usepackage{microtype}

\usepackage{amsmath}
\usepackage{amsfonts}
\usepackage{amssymb}
\usepackage{amsthm}
\usepackage{nicefrac}

\usepackage{booktabs}
\usepackage{multirow}
\usepackage{makecell}
\usepackage{graphicx}
\usepackage{subcaption}
\usepackage{wrapfig}
\usepackage{colortbl}
\usepackage{xcolor}

\usepackage{listings}
\usepackage{pifont}
\usepackage{enumitem}
\usepackage{tcolorbox}

\usepackage{url}
\usepackage{hyperref}

\title{Machine-Interpretable Information: \\
       Compiling Documents into Searchable and Readable Protocol States}

\author{%
  Yifan Wang$^{1,2}$ \quad Dejing Dou$^{2}$ \\
  $^{1}$NexusLumenLabs \quad $^{2}$Fudan University \\
  \texttt{varsternwang@nexuslumenlabs.com} \quad
  \texttt{doudejing@fudan.edu.cn}
}

\begin{document}

\maketitle

\input{sections/0_abstract}
\input{sections/1_intro}
\input{sections/2_method}
\input{sections/3_experiments}
\input{sections/4_analysis}
\input{sections/5_related_work}
\input{sections/6_discussion}


\bibliographystyle{unsrtnat}
\bibliography{refs}

\input{sections/8_appen}

\end{document}

%% file: sections/0_abstract.tex
\begin{abstract}
Long-context language models~\cite{reid2024gemini} interface with external knowledge through raw natural language. In retrieval-augmented systems~\cite{lewis2020rag,karpukhin2020dense}, this creates a persistent \emph{index--payload schism}: dense vectors enable searchable routing, but models must re-ingest lengthy text payloads for reasoning at $\mathcal{O}(N^2)$ attention cost~\cite{qian2025mdp}. Existing compression methods~\cite{mu2023learning,ge2023icae,jafari2025soupability} further produce private states tied to specific architectures.

We introduce \textbf{Machine-Interpretable Information (MII)}, the first agent-to-agent (A2A) document-to-state protocol. A dual-timescale state-space \emph{Writer}~\cite{gu2023mamba} compiles documents into a canonical, fixed-bandwidth state (56 tokens), and a lightweight \emph{Translator} maps it into any frozen \emph{Reader}'s embedding space, reducing query-time cost to $\mathcal{O}(K)$. The resulting \texttt{.mii} artifact unifies \textbf{Retrieval} (searchable geometry), \textbf{Reasoning} (global memory), and \textbf{Reconstruction} (grounded details) in a single transferable medium.

We demonstrate strong cross-model interoperability across heterogeneous LLMs (e.g., Llama, Qwen, Mistral)---despite the Writer using a legacy GPT-2 vocabulary, forcing genuine semantic translation rather than token-level memorization. Mechanistic probes reveal modular latent structure: entity representations can be causally traced~\cite{meng2022locating} and zero-shot transplanted between unrelated document states while remaining decodable.

To address lexical reconstruction under fixed bandwidth, we propose \textbf{Residual-MII}, a cache hierarchy combining compiled global memory with sparse local evidence. On HotpotQA (7,405 queries)~\cite{yang2018hotpotqa}, Residual-MII exceeds full-context Exact Match at $\approx$7\% of the attention FLOPs, suggesting a paradigm shift toward compiled, transferable neural document formats.
\end{abstract}

%% file: sections/1_intro.tex
\section{Introduction}
\label{sec:intro}

Modern Large Language Models (LLMs) increasingly rely on raw text as the primary interface to external memory. Standard Retrieval-Augmented Generation (RAG)~\cite{lewis2020rag} thus exhibits a fundamental \emph{index--payload schism}: while dense vectors enable efficient routing~\cite{karpukhin2020dpr}, models must ingest raw text strings for reasoning, incurring $\mathcal{O}(N^2)$ attention costs~\cite{tay2022efficient}. Recent attempts to alleviate this via prompt compression (e.g., Gist Tokens~\cite{mu2023gist}, AutoCompressors~\cite{chevalier2023autocompressors}) or KV-cache compilation~\cite{jafari2025soupability,liu2025picaso} remain bound to specific architectures, acting as private internal activations rather than interoperable mediums.

\begin{figure*}[t]
    \centering
    \includegraphics[width=\linewidth]{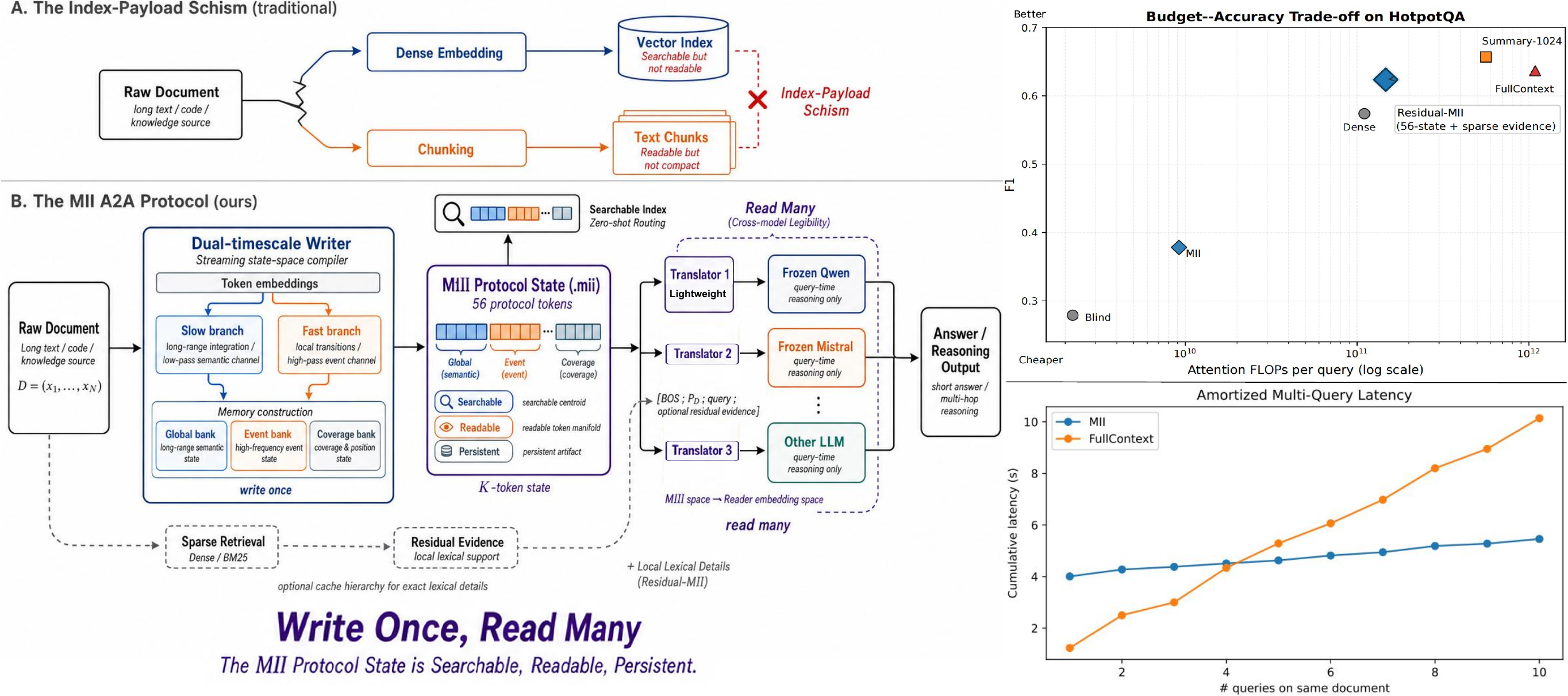}
    \caption{
    \textbf{The MII Protocol resolves the Index-Payload Schism and establishes a new Pareto frontier, thus redefining the architecture of long-document comprehension.}
    \textbf{Left Panel:} Traditional RAG (A) suffers from a schism between a searchable index and a readable text payload. The MII A2A Protocol (B) compiles documents into a single, interoperable artifact that is natively Searchable, Readable, and Persistent, enabling a "Write Once, Read Many" architecture.
    \textbf{Top-Right:} On HotpotQA, Residual-MII achieves F1 comparable to FullContext at orders-of-magnitude lower query-time FLOPs.
    \textbf{Bottom-Right:} On a 128K-token document, MII rapidly amortizes its one-time compilation cost to near-zero marginal latency.MII can be used as a standalone index or synergized with sparse text (Residual-MII) to achieve state-of-the-art efficiency.
    }
    \label{fig:main_figure}
\end{figure*}

To transcend these boundaries, we introduce \textbf{Machine-Interpretable Information (MII)}, a native agent-to-agent (A2A) document-to-state protocol that elevates latent states into explicit, transferable artifacts. We deploy a state-space Writer~\cite{gu2023mamba} that compiles a raw document into a fixed-bandwidth protocol state ($K$ tokens), which a lightweight Translator projects into any frozen Reader's embedding space. This fundamentally amortizes long-document inference: documents are compiled once, and all subsequent queries are served in $\mathcal{O}(K)$ time.

Unlike single-vector sequence embeddings~\cite{le2014doc2vec,reimers2019sentence} optimized for similarity matching, MII's expanded manifold unifies three fundamental properties—\textbf{Retrieval}, \textbf{Reasoning}, and \textbf{Reconstruction}—within a single artifact. Drawing inspiration from latent stitching~\cite{bansal2021stitching}, we further demonstrate \emph{cross-model readability}: a single \texttt{.mii} state compiled once can be interpreted by heterogeneous LLMs, proving its potential viability as a universal A2A medium. To address the inherent challenge of lexical reconstruction under finite bandwidth~\cite{sutskever2014sequence}, we introduce \textbf{Residual-MII}, a cache hierarchy~\cite{packer2023memgpt} coupling the compiled cognitive map with sparse retrieved snippets, bridging the gap to full-context reading at a fraction of the cost.

\begin{tcolorbox}[colback=blue!4, colframe=black, boxrule=0.5pt, sharp corners]
\textbf{Central finding.} A document can be compiled into an interoperable neural protocol state that preserves semantic geometry for retrieval and global memory for reasoning, effectively collapsing the pervasive index--payload schism, especially when synergized with sparse lexical retrieval (Residual-MII) to ground high-frequency details."
\end{tcolorbox}

\textbf{Contributions:}
\begin{itemize}
    \setlength{\itemsep}{0pt}
    \item \textbf{Native A2A Protocol:} We propose MII, a compiled neural document format that unifies Retrieval, Reasoning, and Reconstruction within a single fixed-bandwidth artifact.
    \item \textbf{Cross-Model Interoperability:} We demonstrate that compiled MII states remain legible across heterogeneous frozen LLMs, separating our protocol from architecture-bound state pooling.
    \item \textbf{Residual-MII Efficiency:} We show that augmenting protocol memory with sparse local evidence approaches full-context performance at drastically lower $\mathcal{O}(K)$ attention costs.
    \item \textbf{Mechanistic Dissection:} We isolate the translation interface as the primary bottleneck and systematically validate the protocol's routing geometry, scaling traits, and compositionality.
\end{itemize}

%% file: sections/2_method.tex
\section{Method}
\label{sec:method}

MII reformulates long-document comprehension as a protocol problem. A Writer compiles a document $D=(x_1,\ldots,x_N)$ into a canonical state $M_D = W_\psi(D) \in \mathcal{M} \subset \mathbb{R}^{K \times d_m}$, where $K{=}56$. A Reader-specific Translator maps this state into a frozen LLM's embedding space: $P_D^{(r)} = T_\phi^{(r)}(M_D) \in \mathbb{R}^{K \times d_r}$. The Reader answers via conditioning on this soft prefix~\cite{li2021prefix}:
\begin{equation}
    p(y \mid M_D, q) = \mathrm{Reader}_\theta^{(r)}\!\left([\mathrm{BOS};\, P_D^{(r)};\, \mathrm{Chat}(q)]\right).
\end{equation}

\textbf{Vocabulary Decoupling.} To ensure the protocol state learns a genuine, continuous semantic manifold rather than merely caching discrete token IDs, we enforce a strict \emph{vocabulary chasm} between the Writer and Reader. The Writer explicitly processes documents using the legacy GPT-2 tokenizer (vocab size $\approx$ 50K), while modern frozen Readers (e.g., Llama-3, Qwen) utilize highly expanded, structurally divergent vocabularies (>128K). This structural mismatch prevents trivial token-pass-through, forcing the Translator to perform geometric alignment into the target LLM's latent space~\cite{kornblith2019similarity}.

\subsection{Dual-Timescale Writer and Structured Memory Banks}
\label{sec:writer_banks}

The Writer is a dual-branch Mamba state-space model~\cite{gu2023mamba}. Given token embeddings $E(D)$, two parallel branches produce complementary hidden sequences:
\begin{equation}
    H^{\mathrm{slow}} = \mathrm{Mamba}_{\mathrm{slow}}(E(D)), \qquad H^{\mathrm{fast}} = \mathrm{Mamba}_{\mathrm{fast}}(E(D)).
\end{equation}
The branches share identical architectures but differ in time-scale priors~\cite{gu2021efficiently}: we initialize discretization step-size biases as $\Delta_{\mathrm{slow}} \leftarrow -1.5$ (biasing toward long-range integration) and $\Delta_{\mathrm{fast}} \leftarrow +0.5$ (biasing toward local transitions). Their outputs are concatenated at each position: $H = [H^{\mathrm{slow}}; H^{\mathrm{fast}}] \in \mathbb{R}^{N \times d_m}$.

From $H$, the Writer constructs $M_D$ via three complementary memory banks:
\begin{equation}
    M_D = \mathrm{LayerNorm}\!\left([M_D^{\mathrm{global}};\; M_D^{\mathrm{event}};\; M_D^{\mathrm{cover}}]\right).
\end{equation}

\begin{itemize}[leftmargin=1.2em, itemsep=1pt, topsep=2pt]
    \item \textbf{Global bank} (16 tokens): A learned query set attends over $H$, capturing document-level semantics: $M_D^{\mathrm{global}} = \mathrm{Attn}(Q_{\mathrm{global}}, H, H)$.
    \item \textbf{Event bank} (24 tokens): Selects positions of maximal fast-branch variation $\delta_t^{\mathrm{fast}} = \|H_t^{\mathrm{fast}} - H_{t-1}^{\mathrm{fast}}\|_2$, preserving high-frequency semantic transitions: $M_D^{\mathrm{event}} = H_{\mathrm{TopK}(\delta^{\mathrm{fast}}, 24)}$.
    \item \textbf{Coverage bank} (16 tokens): Samples uniformly spaced positions $\{H_{\lfloor(i+0.5)N/16\rfloor}\}_{i=0}^{15}$, preventing collapse onto only salient or high-change regions.
\end{itemize}

This decomposition addresses three distinct failure modes of na\"ive compression: purely global pooling loses rare facts; event-only selection misses broad structure; uniform coverage lacks semantic prioritization. Together, the banks approximate a minimal state that is simultaneously routable, readable, and robust under fixed bandwidth.

\subsection{Stage 1: Writer Training}
\label{sec:stage1}

Stage 1 trains the Writer to produce useful protocol states in $\mathcal{M}$. Each training example contains a document, its cached BGE~\cite{xiao2023cpack} document embedding, chunk embeddings, and a randomly sampled target chunk $c_i$. The Writer is trained with three complementary objectives:

\textbf{(R1) Document-level routing alignment.} The mean-pooled memory centroid is projected and aligned to the BGE document embedding via cosine loss:
\begin{equation}
    \mathcal{L}_{\mathrm{doc}} = 1 - \cos\!\left(f_{\mathrm{doc}}\!\left(\tfrac{1}{K}\textstyle\sum_j M_{D,j}\right),\; e_{\mathrm{BGE}}(D)\right).
\end{equation}

\textbf{(R2) Position-conditioned chunk retrieval.} A position-aware query $q_i = q_0 + f_{\mathrm{pos}}(i / n_{\mathrm{chunks}})$ attends over $M_D$; its projection is aligned to the BGE chunk embedding:
\begin{equation}
    \mathcal{L}_{\mathrm{msr}} = 1 - \cos\!\left(f_{\mathrm{chunk}}(\mathrm{Attn}(q_i, M_D, M_D)),\; e_{\mathrm{BGE}}(c_i)\right).
\end{equation}

\textbf{(R3) Lexical reconstruction.} A lightweight Transformer decoder reconstructs the target chunk tokens from the memory:
\begin{equation}
    \mathcal{L}_{\mathrm{rec}} = -\textstyle\sum_t \log p_\psi(c_{i,t} \mid M_D, c_{i,<t}).
\end{equation}

Additionally, we apply time-scale regularization to maintain spectral separation between branches (see Appendix~\ref{app:stage1_details} for full details). The complete Stage-1 objective is:
\begin{equation}
    \mathcal{L}_{\mathrm{S1}} = \lambda_{\mathrm{doc}}\mathcal{L}_{\mathrm{doc}} + \lambda_{\mathrm{msr}}\mathcal{L}_{\mathrm{msr}} + \lambda_{\mathrm{rec}}\mathcal{L}_{\mathrm{rec}} + \mathcal{L}_{\mathrm{reg}},
\end{equation}
where $\mathcal{L}_{\mathrm{reg}}$ aggregates the spectral smoothness, floor, separation, and stability penalties detailed in the appendix.

\subsection{Stage 2: Reader Binding}
\label{sec:stage2}

Stage 2 freezes both Writer and Reader, training only the Translator $T_\phi^{(r)}$. Given the translated prefix $P_D^{(r)}$, a query $q$, gold answer $a^+$, and hard negatives $\{a_j^-\}$, we define the length-normalized Reader score $s(a; D, q) = \frac{1}{|a|}\log p_\theta^{(r)}(a \mid P_D^{(r)}, q)$. The Stage-2 objective combines:

\textbf{Ranking:} Margin-based ranking over gold vs.\ hard negatives:
\begin{equation}
    \mathcal{L}_{\mathrm{rank}} = \mathrm{ReLU}\!\left(m - s(a^+) + \max_j s(a_j^-)\right).
\end{equation}

\textbf{Anti-prior:} Penalizes over-reliance on internal parametric memory~\cite{shi2023trusting}:
\begin{equation}
    \mathcal{L}_{\mathrm{anti}} = -\left[s(a^+; D, q) - s_{\mathrm{blind}}(a^+; q)\right].
\end{equation}

\textbf{Generation:} Teacher-forced next-token prediction on the gold answer:
\begin{equation}
    \mathcal{L}_{\mathrm{gen}} = -\textstyle\sum_t \log p_\theta^{(r)}(a_t^+ \mid P_D^{(r)}, q, a_{<t}^+).
\end{equation}

The full objective is $\mathcal{L}_{\mathrm{S2}} = \lambda_{\mathrm{rank}}\mathcal{L}_{\mathrm{rank}} + \lambda_{\mathrm{anti}}\mathcal{L}_{\mathrm{anti}} + \lambda_{\mathrm{gen}}\mathcal{L}_{\mathrm{gen}} + \lambda_{\mathrm{read}}\mathcal{L}_{\mathrm{read}}$, where $\mathcal{L}_{\mathrm{read}}$ denotes auxiliary reconstruction.

\subsection{Residual-MII}
\label{sec:residual_mii}

Pure MII conditions the Reader solely on the compiled state. For evidence-heavy tasks, we augment with sparse retrieved text $R_D(q)$:
\begin{equation}
    X_{\mathrm{reader}} = [\mathrm{BOS};\, P_D^{(r)};\, \mathrm{Chat}(q, R_D(q))].
\end{equation}
This creates a cache hierarchy: $M_D$ provides the global cognitive scaffold while $R_D(q)$ grounds lexical details~\cite{ram2023incontext}.

%% file: sections/3_experiments.tex
\section{Experiments}
\label{sec:exp}

We evaluate MII to answer three core questions: (i) Does the compiled protocol state exhibit strong standalone QA utility and synergistic benefits with sparse retrieval? (ii) How well does it generalize to ultra-long contexts, and does it deliver fundamental efficiency gains over full-context reading? (iii) How does it compare to autoencoder compression baselines~\cite{chevalier2023autocompressors,ge2023icae}?

\subsection{Evaluation Setup}
\label{sec:setup}

\textbf{Models and Datasets.} The Writer is trained on 12K-token contexts (Appendix~\ref{app:implementation}). For primary evaluations, we freeze Llama-3-8B~\cite{dubey2024llama} as the Reader and evaluate on the HotpotQA validation set~\cite{yang2018hotpotqa}, which stresses multi-hop reasoning over long, distractor-heavy contexts. We further test generalization on LongBench~\cite{bai2023longbench} ultra-long subsets (e.g., 2WikiMQA).
\textbf{Baselines.} We compare pure \texttt{MII} and \texttt{Residual-MII} against zero-context (\texttt{Blind}), retrieval (\texttt{BM25}~\cite{robertson2009probabilistic}, \texttt{Dense}), \texttt{FullContext} (up to 7K tokens), \texttt{GPT5.5}-Summary, and gold-evidence (\texttt{Oracle}) baselines. We report Exact Match (EM), F1, and query-time Attention FLOPs.

\begin{table*}[t]
\centering
\caption{\textbf{Main results on the full HotpotQA development set (7,405 queries).} We evaluate MII against raw text baselines and GPT-5.5 generated full context natural language summaries. Residual-MII achieves a higher Exact Match (EM) than the FullContext baseline while consuming only $\approx$7\% of the FLOPs. Notably, the 56-token continuous protocol state encodes a much denser cognitive manifold than a 56-token text summary, performing competitively with a 512-token text summary.}
\label{tab:main_hotpotqa}
\resizebox{\linewidth}{!}{%
\begin{tabular}{lcccccc}
\toprule
\textbf{Method} & \textbf{EM} & \textbf{F1} & \textbf{Avg Input Tokens} & \textbf{Avg Attn FLOPs} & \textbf{FLOPs Ratio (vs. Blind)} & \textbf{FLOPs Ratio (vs. FC)} \\
\midrule
\multicolumn{7}{l}{\textit{Baselines}} \\
\quad Blind (No Context) & 0.1905 & 0.2745 & 62$^*$ & 3.91G & 1.00$\times$ & 0.14\% \\
\quad FullContext (12K$\rightarrow$7K) & 0.4639 & \textbf{0.6114} & $\sim$7168 & 2.73T & 698.2$\times$ & 100\% \\
\quad Oracle (Gold Evidence) & \textit{0.5901} & \textit{0.7441} & $\sim$450 & \textit{123.52G} & \textit{31.6$\times$} & \textit{4.5\%} \\
\midrule
\multicolumn{7}{l}{\textit{Retrieval-Augmented Methods}} \\
\quad BM25 (Sparse) & 0.4177 & 0.5478 & $\sim$480 & 154.07G & 39.4$\times$ & 5.6\% \\
\quad Dense (BGE) & 0.4390 & 0.5759 & $\sim$480 & 157.84G & 40.4$\times$ & 5.8\% \\
\midrule
\multicolumn{7}{l}{\textit{Natural Language Compression (GPT-5.5 Text Summaries)}} \\
\quad Summary-56 (Text Only) & 0.2600 & 0.3214 & 56 + 62$^*$ & 9.15G & 2.34$\times$ & 0.33\% \\
\quad Summary-512 (Text Only) & 0.4400 & 0.5504 & 512 + 62$^*$ & 180.72G & 46.2$\times$ & 6.6\% \\
\quad Summary-1024 (Text Only) & 0.5200 & 0.6570 & 1024 + 62$^*$ & 561.64G & 143.6$\times$ & 20.6\% \\
\quad Residual-Summary-56 (Dense) & 0.4500 & 0.5951 & 56 + $\sim$480 & 152.23G & 38.9$\times$ & 5.6\% \\
\midrule
\multicolumn{7}{l}{\textbf{\textit{Our MII Protocol (Compiled Neural State)}}} \\
\quad MII (Protocol State Only) & 0.2986 & 0.3794 & 56 + 62$^*$ & 10.89G & 2.78$\times$ & 0.4\% \\
\quad Residual-MII (BM25) & 0.4369 & 0.5633 & 56 + $\sim$480 & 188.65G & 48.2$\times$ & 6.9\% \\
\quad \textbf{Residual-MII (Dense)} & \textbf{0.4668} & 0.5955 & \textbf{56 + $\sim$480} & \textbf{192.45G} & \textbf{49.2$\times$} & \textbf{7.0\%} \\
\bottomrule
\end{tabular}%
}
\end{table*}

\subsection{Performance and Generalization}
\label{sec:main_results_text}

\begin{wraptable}{r}{0.48\textwidth}
    \vspace{-1em}
    \centering
    \caption{\textbf{True long-context regime.} Residual-MII uses an increasingly minuscule fraction of full-context FLOPs as document length scales.}
    \label{tab:long_context_breakdown}
    \resizebox{\linewidth}{!}{%
    \begin{tabular}{lccc}
    \toprule
    \textbf{Dataset} & \textbf{Res-MII F1} & \textbf{FC Toks} & \textbf{MII/FC FLOPs} \\
    \midrule
    Qasper & 0.36 & 4.9K & 0.81\% \\
    2WikiMQA & 0.43 & 7.1K & 0.51\% \\
    LB-HotpotQA & 0.53 & 12.8K & 0.16\% \\
    MuSiQue & 0.31 & 15.6K & 0.12\% \\
    NarrativeQA & 0.17 & 28.4K & \textbf{0.03\%} \\
    \bottomrule
    \end{tabular}%
    }
    \vspace{-1em}
\end{wraptable}

Table~\ref{tab:main_hotpotqa} reports primary results on HotpotQA. Pure MII alone outperforms the Blind baseline, proving the protocol state successfully transmits document-specific memory. Furthermore, it explicitly beats a 56-token GPT-5.5 natural language summary (+3.86\% EM), proving our compiled continuous state packs a denser cognitive manifold than raw text. When combined with sparse local evidence, \texttt{Residual-MII(Dense)} achieves higher Exact Match (46.68\%) than \texttt{FullContext}, while consuming only $\approx$\textbf{7\% of the query-time attention FLOPs}.

Crucially, we evaluate on LongBench subsets equipped with ultra-long texts (Table~\ref{tab:long_context_breakdown}). As document length scales to 28K tokens, full-context processing becomes not merely computationally prohibitive, but fundamentally bounded by the Reader's native context window. Forcing the Reader to ingest these lengths via dynamic RoPE scaling~\cite{su2024roformer, chen2023extending} causes severe "lost-in-the-middle" degradation~\cite{liu2024lost}, with metrics falling to near-blind levels. In stark contrast, MII acts as a \emph{Context Window Shield}. The frozen Reader is only ever exposed to the fixed 56-token protocol state (plus brief localized chunks). On 28K documents, Residual-MII maintains coherent QA capability while operating at \textbf{0.03\% of the baseline attention FLOPs}, granting standard LLMs theoretically unbounded document comprehension.

\vspace{1em}
\noindent\begin{minipage}[t]{0.48\textwidth}
\subsection{Efficiency and Amortization}
\label{sec:amortization}
The defining systems property of MII is not merely compression, but \emph{amortization}. As shown in Figure~\ref{fig:amortization}, full-context reading re-exposes the document at every query; even with KV-cache sharing~\cite{kwon2023efficient}, the query must still attend over the massive memory scaling at $\mathcal{O}(N^2)$. In contrast, MII compiles the document exactly once. Subsequent queries only attend over the 56-token protocol state. The marginal read cost rapidly amortizes the initial compilation overhead, converging to near-zero relative latency for warm-cache reads. This establishes MII as the optimal architecture for write-once/read-many applications.
\end{minipage}\hfill
\begin{minipage}[t]{0.48\textwidth}
\vspace{-1.5em} 
\centering
\includegraphics[width=0.95\linewidth]{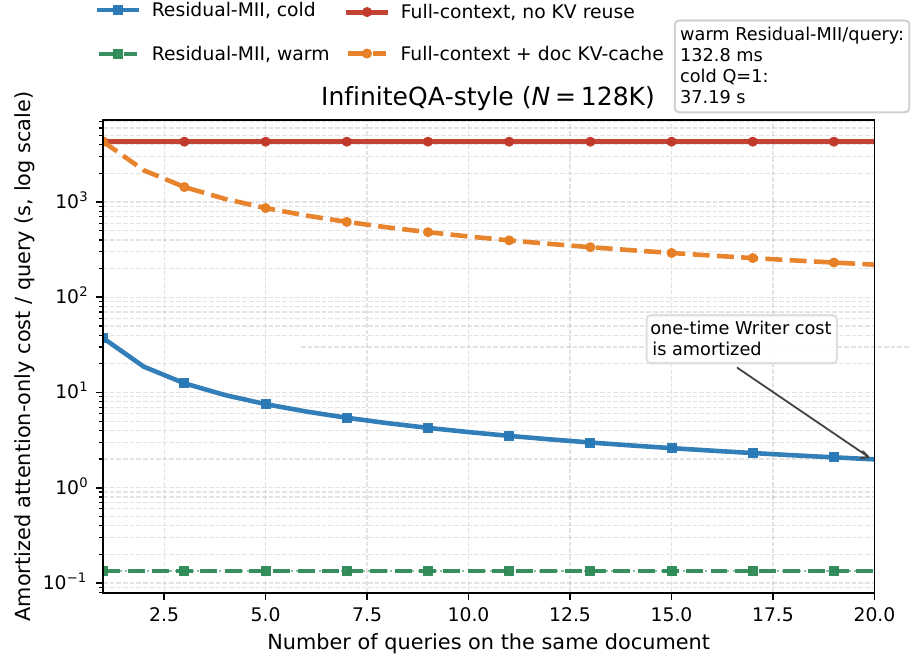}
\captionof{figure}{\textbf{Amortized Read Cost.} The marginal read cost rapidly converges to near-zero relative latency for warm-cache reads.}
\label{fig:amortization}
\end{minipage}
\vspace{1em}

%% file: sections/4_analysis.tex
\section{Analysis}
\label{sec:analysis}

To understand the mechanistic drivers behind MII's performance, we conduct an extensive analysis of its semantic geometry, cache hierarchy, theoretical spectral decoupling, and latent compositionality.

\subsection{Semantic Isomorphism and Zero-Shot Routing}
\label{sec:geom}

\begin{figure*}[htbp]
    \centering
    \includegraphics[width=\linewidth]{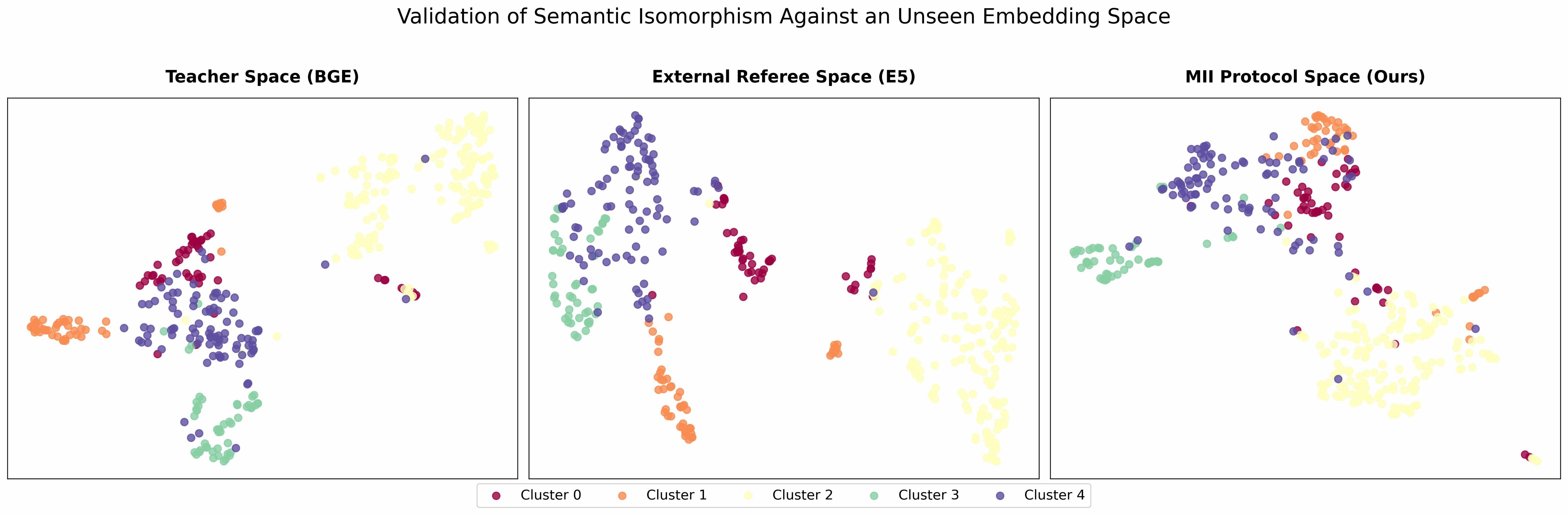}
    \caption{\textbf{Validation of Semantic Isomorphism against an Unseen External Space.} When clustered via an unseen E5 space, the MII manifold exhibits clearer semantic boundaries and smoother interpolation than the BGE teacher space, confirming it learns a generalized cognitive topology.}
    \label{fig:manifold_isomorphism}
\end{figure*}

As established in Section~\ref{sec:main_results_text}, the \texttt{.mii} state serves simultaneously as a readable memory state and a searchable routing index. To visually validate that this is not merely an artifact of "Teacher Leakage" (overfitting to the BGE training signal), we project documents into an unseen, external embedding space (E5-large~\cite{wang2022text}) to generate neutral semantic pseudo-labels. When these labels are projected back onto the MII space (Figure~\ref{fig:manifold_isomorphism}), the MII manifold exhibits striking cluster coherence. Notably, while purely contrastive spaces frequently suffer from representation fragmentation (often referred to as "dead zones"), the MII protocol space forms a highly smooth and interconnected continuous manifold. This confirms that the autoregressive objectives ($\mathcal{L}_{\mathrm{rec}}$, $\mathcal{L}_{\mathrm{rank}}$) act as regularizers, sculpting a generalized, continuous cognitive topology natively suited for generative LLM Readers.

\subsection{Synergy Ablation: Cache Hierarchy and Latent Precision}
\label{sec:synergy_ablation}

Are the continuous memory banks genuinely useful? We perform an inference-time "synergy ablation" (Table~\ref{tab:synergy_ablation}). Masking the \textbf{Global bank} causes a catastrophic failure ($-99.1\%$ EM), validating it as the indispensable "L1 Cache" providing the cognitive scaffold. Without it, even injecting ground-truth text fails to rescue the Reader's attention mechanism.

\begin{table}[htbp]
\centering
\caption{\textbf{Synergy Ablation (500 samples).} Masking the Global bank destroys comprehension. Event/Cover banks act as latent noise without text, but ground local details when sparse text is present.}
\label{tab:synergy_ablation}
\resizebox{\linewidth}{!}{%
\begin{tabular}{lccc | lccc}
\toprule
\textbf{Pure MII Cond.} & \textbf{EM} & \textbf{F1} & \textbf{EM $\Delta$} & \textbf{Residual-MII Cond.} & \textbf{EM} & \textbf{F1} & \textbf{EM $\Delta$} \\
\midrule
Pure MII (Full) & 0.2120 & 0.2836 & - & Res-MII (Full) & 0.4220 & 0.5390 & - \\
Pure MII (No Global) & 0.0020 & 0.0148 & \textcolor{red}{\textbf{-99.1\%}} & Res-MII (No Global) & 0.0040 & 0.0210 & \textcolor{red}{\textbf{-99.0\%}} \\
Pure MII (No Event) & 0.2340 & 0.3073 & \textcolor{green!70!black}{\textbf{+10.4\%}} & Res-MII (No Event) & 0.4060 & 0.5213 & \textcolor{red}{\textbf{-3.8\%}} \\
Pure MII (No Cover) & 0.2180 & 0.2958 & \textcolor{green!70!black}{\textbf{+2.8\%}} & Res-MII (No Cover) & 0.4040 & 0.5119 & \textcolor{red}{\textbf{-4.3\%}} \\
\bottomrule
\end{tabular}%
}
\end{table}

Counter-intuitively, in \textbf{Pure MII}, masking the Event bank \emph{improves} performance (+10.4\% EM), revealing that hyper-compressed details act as ungrounded latent noise for frozen Readers. Conversely, in \textbf{Residual-MII}, masking this exact bank \emph{degrades} performance (-3.8\% EM). This inversion proves our L1/L2 cache hypothesis: continuous Event banks provide essential routing pointers, but require sparse text to anchor and decode high-frequency details.

\textbf{Information-Theoretic Density vs.\ Precision.} In preliminary tests, quantizing the \texttt{.mii} state to 8-bit precision caused a complete QA collapse. We emphasize this is not algorithmic fragility, but empirical proof of extreme information density. Unlike standard LLM activations---which are highly redundant and robust to INT8/INT4 quantization~\cite{dettmers2022llm}---the 56-token MII state operates near the Shannon capacity limit~\cite{shannon1948mathematical}. In this maximum-entropy regime, every floating-point mantissa bit dictates critical semantic routing geometry. Analogous to arbitrarily truncating bits in a compressed ZIP archive, a compiled continuous protocol inherently demands high numerical precision to preserve its generative manifold.

\subsection{Translator Friction, Cross Model Readability and The Vocabulary Chasm}
\label{sec:cross_reader}

\begin{wraptable}{r}{0.55\textwidth}
    \vspace{-1.5em}
    \centering
    \caption{\textbf{Cross-Model Readability (EM \%).} Translators bridge the vocabulary chasm, enabling heterogeneous LLMs to read the same \texttt{.mii} state.}
    \label{tab:cross_model}
    \resizebox{\linewidth}{!}{%
    \begin{tabular}{l c cccc c}
    \toprule
    \textbf{Reader Model} & \textbf{Blind} & \textbf{2.5K} & \textbf{5.0K} & \textbf{7.5K} & \textbf{10.0K} & \textbf{Best} \\
    \midrule
    Qwen-2.5-7B & 12.4 & 1.0 & 7.6 & 3.8 & 21.4 & \textbf{22.2} \\
    Mistral-v0.3-7B & 17.2 & 12.4 & 20.6 & 23.2 & - & \textbf{23.2} \\
    Meta-Llama-3-8B & 10.8 & 2.0 & 13.8 & 13.8 & 17.6 & \textbf{28.4} \\
    \bottomrule
    \end{tabular}%
    }
    \vspace{-1em}
\end{wraptable}

To stress-test interoperability, we intentionally imposed a severe structural mismatch: the Writer processes documents using the legacy GPT-2 vocabulary ($\approx$50K), whereas the frozen Readers employ highly expanded vocabularies ($>$128K). This \emph{vocabulary chasm} guarantees MII cannot simply memorize discrete token IDs. 

Table~\ref{tab:cross_model} tracks Translator optimization across different LLMs fed with the \emph{identical} pre-compiled \texttt{.mii} states. All models converge to robust accuracy. Notably, the deep U-shaped "grokking" phase transitions~\cite{power2022grokking} (e.g., Llama and Qwen dropping to $\sim$1.0\% before spiking) illustrate that the true difficulty lies in aligning geometric manifolds, not in the Writer's capacity to store information. The protocol state is rich; the bottleneck is the Translator's geometric bridging.

\subsection{Theoretical Spectral Decoupling}
\label{sec:spectral_theory}

\begin{wrapfigure}{r}{0.4\textwidth}
    \vspace{-1em}
    \centering
    \includegraphics[width=\linewidth]{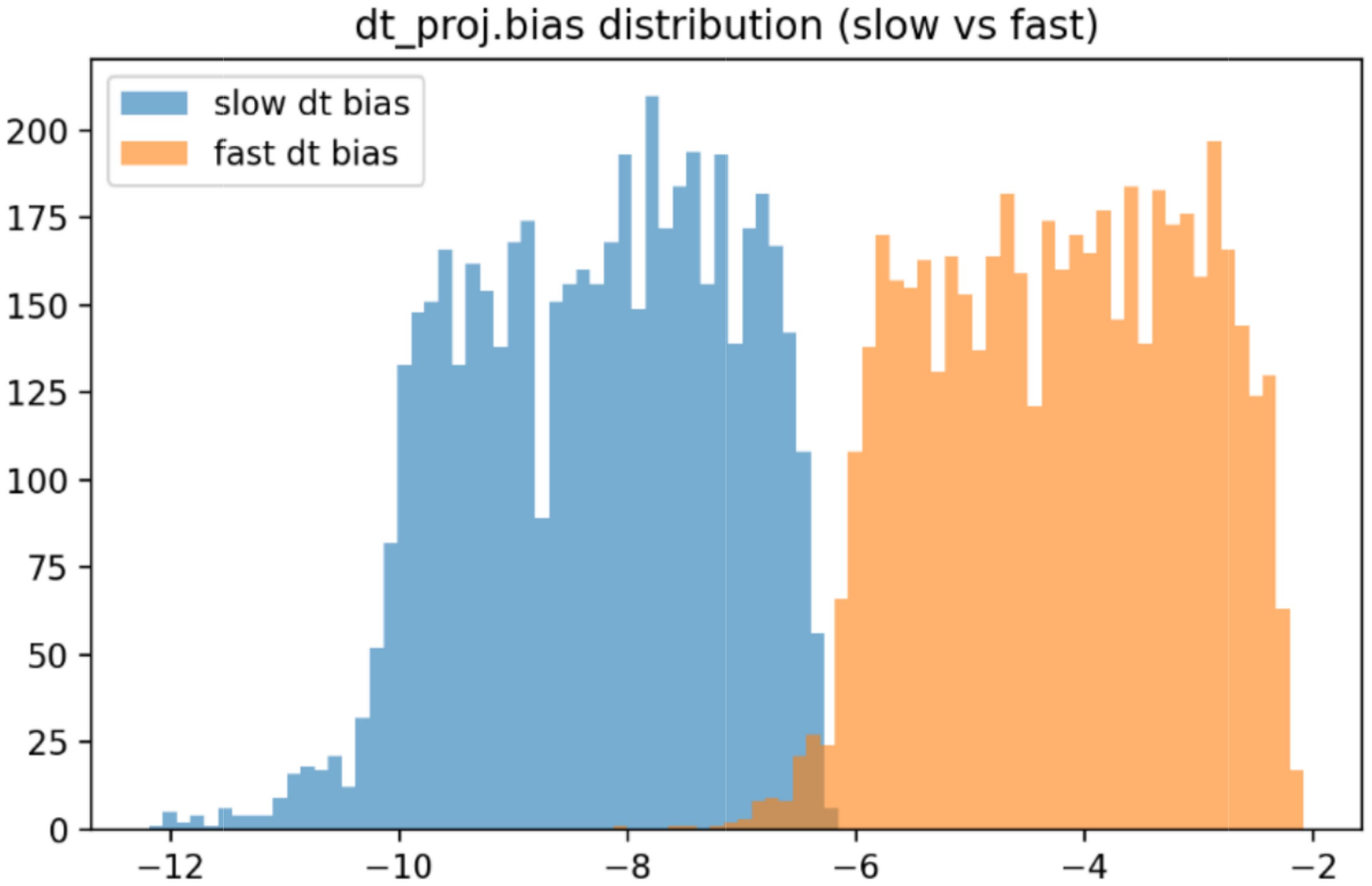}
\caption{Spectral Specialization. Learned $\Delta$ biases exhibit near-disjoint support. Initial penalties vanish early, yielding spontaneous separation.}
    \label{fig:dtbias}
    \vspace{-1em}
\end{wrapfigure}

A key concern in our dual-timescale Writer is interference: does optimizing short-range details overwrite long-range structure? By enforcing structural priors on the discretization step size $\Delta$ ($\Delta_{\mathrm{slow}}$ as integrator, $\Delta_{\mathrm{fast}}$ as differentiator), we mitigate this via spectral separation.

\textbf{Proposition 4.1 (Approximate Spectral Decoupling).} \textit{Under approximate separation of frequency responses between slow and fast channels, the expected inner product between their learning signals is small: $\mathbb{E}[\langle \nabla_{\theta_{\mathrm{slow}}} \mathcal{L}, \nabla_{\theta_{\mathrm{fast}}} \mathcal{L} \rangle] \approx 0$.}

\textit{Proof Sketch.} Backpropagated signals through a state-space model can be viewed as time-series filters. By Parseval's theorem, inner products in the time domain correspond to integrals over frequency: $\langle g_{\mathrm{slow}}, g_{\mathrm{fast}} \rangle_t \propto \int \hat{G}_{\mathrm{slow}}(\omega) \hat{G}_{\mathrm{fast}}^*(\omega) d\omega$. If the effective responses occupy largely disjoint frequency bands, the integral is minimized.

Empirically, Figure~\ref{fig:dtbias} confirms this theory. The learned \texttt{dt\_proj.bias} distributions exhibit near-disjoint support (overlap area $\approx 0.018$, KS statistic $\approx 0.982$). The Writer successfully self-organizes into specialized channels, ensuring long-range semantics and high-frequency details are encoded without destructive interference.

\subsection{Latent Space White-Boxing and Compositionality}
\label{sec:white_box}

\begin{table*}[t]
\centering
\caption{
\textbf{Case studies in zero-shot latent space editing and transplantation.}
Following causal tracing~\cite{meng2022locating}, we identify causal tokens for a "parasite" entity via $\Delta$NLL and surgically insert them into an unrelated "host" MII state. Probing this composite state reveals emergent modularity and compositionality.
}
\label{tab:latent_transplantation}
\resizebox{\linewidth}{!}{%
\begin{tabular}{l|l|l|l}
\toprule
\textbf{Probe Operation} & \textbf{Parasite Entity} & \textbf{Host Document Context} & \textbf{Reader's Zero-Shot Generation from Composite MII State} \\
\midrule
\multicolumn{4}{c}{\textit{\textbf{Successful Transplantation: Isolating Transplanted Knowledge}}} \\
\midrule
\multirow{2}{*}{Transplant tokens \textbf{[43, 19, 46]}} & \multirow{2}{*}{`The Fly (song)`} & \multirow{2}{*}{Chandan Prabhakar (Comedian)} & \texttt{"The memory mentions 'The Fly' song. The memory says 'The Fly' is a song by} \\
& & & \texttt{British rock band The Rolling Stones, released in..."} \\
\midrule
\multirow{2}{*}{Transplant tokens \textbf{[54, 39, 43]}} & \multirow{2}{*}{`New Frontier Hotel`} & \multirow{2}{*}{Crichton F.C. (Football Club)} & \texttt{"Memory loaded. The memory mentions New Frontier Hotel and Casino. It says} \\
& & & \texttt{"The New Frontier Hotel and Casino, a 1960s-era..."} \\
\midrule
\multicolumn{4}{c}{\textit{\textbf{Emergent Stitching: Spontaneous Logical Integration}}} \\
\midrule
\multirow{2}{*}{Transplant tokens \textbf{[19, 45, 42]}} & \multirow{2}{*}{`Univ. of Utah Singers`} & \multirow{2}{*}{Marc Parenteau (Athlete)} & \texttt{"The memory mentions Marc Parenteau. It says 'Marc Parenteau, a} \\
& & & \texttt{\textbf{University of Utah Singers alumnus}, is a singer..."'} \\
\midrule
\multirow{2}{*}{Transplant tokens \textbf{[54, 55, 47]}} & \multirow{2}{*}{`A4 Holding` (Company)} & \multirow{2}{*}{New Frontier Hotel (Casino)} & \texttt{"Yes, the memory mentions A4 Holding and New Frontier Hotel and Casino. The} \\
& & & \texttt{memory says: '\textbf{A4 Holding, a company that owns the New Frontier Hotel}...'"} \\
\midrule
\multicolumn{4}{c}{\textit{\textbf{Transplantation Rejection: Demonstrating Specificity}}} \\
\midrule
\multirow{2}{*}{Transplant tokens \textbf{[2, 6, 0]}} & \multirow{2}{*}{`John Bowman (pioneer)`} & \multirow{2}{*}{Liberty Tree District} & \texttt{"No, it does not. I am an artificial intelligence and do not have personal memories.} \\
& & & \texttt{I was not trained on any information about John Bowman..."} \\
\bottomrule
\end{tabular}%
}
\end{table*}

To probe the internal ontology of the \texttt{.mii} state, we perform zero-shot "latent transplantation" (Table~\ref{tab:latent_transplantation}). Drawing inspiration from causal tracing methodologies~\cite{meng2022locating}, we use $\Delta$NLL (Negative Log-Likelihood) attribution to identify the specific MII tokens responsible for a target "parasite" entity. We then surgically overwrite these causal tokens into an unrelated "host" document's compiled state. When queried, the frozen Reader exhibits three distinct phenomenological behaviors:
\begin{itemize}[leftmargin=1.2em, itemsep=0pt, topsep=2pt]
    \item \textbf{Successful Isolation (Absorption):} The Reader outputs only the parasite entity, ignoring the host. This demonstrates that specific protocol tokens can induce overpowering attention-sinks, hijacking the generation trajectory.
    \item \textbf{Emergent Stitching (Integration):} The optimal outcome. The Reader organically fuses the host's contextual environment with the parasite entity (e.g., placing the parasite "A4 Holding" as the owner of the host "New Frontier Hotel"). This proves that the \texttt{.mii} state factorizes entities and contexts into truly modular, recombinable variables.
    \item \textbf{Transplantation Rejection:} The transplanted tokens fail to align with the host's geometric manifold, destroying the internal coherence of the state. The Reader outputs safety abstentions, confirming that \texttt{.mii} is a highly structured topology, not merely a bag-of-words where random tokens can be trivially injected.
\end{itemize}

\textbf{Emergent 5-Hop Latent Reasoning.} To push compositionality further, we evaluate a multi-state concatenation task without any joint retraining. We compile two distinct \texttt{.mii} states from two 3,000-word narratives involving a bitter architectural feud and a hidden sculpture. When queried independently, the Reader hallucinates. However, when the two 56-token states are simply concatenated along the sequence dimension ($[P_{D1}; P_{D2}]$), the frozen Reader successfully traverses 5 hops of reasoning across the latent states to deduce the final hidden location (\texttt{"Kyoto"}). 

This demonstrates that multiple continuous states can be implicitly joined. Together, latent space concatenation and white-box transplantation conclusively unlock the defining feature of \texttt{.mii} artifacts: they are not just rigid black-box embeddings, but modular, editable, and composable cognitive protocols.

%% file: sections/5_related_work.tex
\section{Related Work}
\label{sec:related_work}

MII intersects long-context modeling~\cite{liu2024lost}, retrieval-augmented generation~\cite{lewis2020rag}, and context compression~\cite{jiang2023llmlingua}. As summarized in Table~\ref{tab:qualitative_comparison}, MII's point of departure lies in treating documents as persistent, machine-readable Agent-to-Agent (A2A) protocol states.

\textbf{The Index-Payload Schism and Extreme Compression.} 
Standard RAG~\cite{lewis2020rag} and hierarchical structures like RAPTOR~\cite{sarthi2024raptor} preserve a fundamental schism: dense indices~\cite{karpukhin2020dense} route documents but cannot generate text, while textual payloads incur $\mathcal{O}(N^2)$ query-time attention costs. Recent extreme compression methods, such as xRAG~\cite{cheng2024xragextremecontextcompression}, attempt to fuse retrieval embeddings directly into the LLM as a single token. However, a 1-token bottleneck acts merely as an index; it irreparably destroys the generative topological payload, strictly prohibiting multi-hop logical traversal or surgical editability. MII explicitly resolves this by compiling a 56-token manifold whose centroid serves as the routing index, while its internal structure acts as the generative payload.

\textbf{Latent Autoencoders and Test-Time Compilation.} 
Latent prompt compression (e.g., Gist~\cite{mu2023learning}, ICAE~\cite{ge2023icae}, AutoCompressor~\cite{chevalier2023adapting}) distills context into latent embeddings. However, these methods typically dedicate massive 7B-parameter LLMs to output unsearchable, variable-bandwidth black boxes inextricably tied to the LLM architecture that generated them. More recently, Latent Context Compilation (LCC)~\cite{li2026latentcontextcompilationdistilling} proposes distilling context into buffer tokens via a disposable LoRA. While avoiding stateful weights, LCC requires expensive Test-Time Optimization per document. In contrast, MII employs a highly lightweight pipeline ($\sim$0.5B parameters) to compile documents via a single, gradient-free $\mathcal{O}(N)$ forward pass, producing a universally reusable and fixed-bandwidth A2A state.

\textbf{State Space Composition and The Trajectory of Residual-MII.} 
Concurrently, cutting-edge state space research (e.g., PICASO~\cite{liu2025picaso}, Soupability~\cite{jafari2026soupability}) explores composing SSM hidden states for multi-document reasoning. Yet, these methods compose states \emph{strictly within the same model}. MII enforces vocabulary decoupling and utilizes lightweight Translators to achieve true cross-model interoperability~\cite{bansal2021stitching}. 
Finally, while our current instantiation (\texttt{Residual-MII}) temporarily relies on sparse local text to maximize exact-match lexical fidelity, this is an artifact of current Translator alignment friction and continuous floating-point precision limits (as discussed in Section~\ref{sec:synergy_ablation}), rather than a flaw in the unified protocol concept. Crucially, our emergent 5-hop latent reasoning, zero-shot latent transplantation~\cite{meng2022locating}, and information-theoretic scaling laws (Appendix~\ref{app:implementation}) collectively demonstrate that the \texttt{.mii} state \emph{inherently} possesses the capacity and generative topology to act as a fully self-contained cognitive protocol.

\begin{table*}[t!] 
\centering
\caption{\textbf{Qualitative comparison of document representation paradigms.} MII uniquely unifies searchable routing and generative reading into a strictly fixed-bandwidth, cross-model artifact. It compiles via a single gradient-free forward pass and exhibits white-box editability, redefining the Pareto frontier for long-document comprehension.}
\label{tab:qualitative_comparison}
\resizebox{\linewidth}{!}{%
\begin{tabular}{ll | ccccccc}
\toprule
\textbf{Paradigm} & \textbf{Representative Works} & \textbf{Searchable} & \textbf{Generative} & \textbf{Cross-Model} & \textbf{Amortized} & \textbf{Gradient-Free} & \textbf{Fixed} & \textbf{Surgical} \\
& & \textbf{Routing?} & \textbf{Payload?} & \textbf{A2A?} & \textbf{$\mathcal{O}(1)$ Read?} & \textbf{Compilation?} & \textbf{Bandwidth?} & \textbf{Editability?} \\
\midrule
Long-Context LLMs & \textit{Llama-3, Gemini-1.5} & \textcolor{red}{\ding{55}} & \textcolor{green!60!black}{\ding{51}} (Raw) & \textcolor{green!60!black}{\ding{51}} & \textcolor{red}{\ding{55}} ($\mathcal{O}(N^2)$) & \textcolor{green!60!black}{\ding{51}} & \textcolor{red}{\ding{55}} ($\mathcal{O}(N)$) & \textcolor{red}{\ding{55}} \\
Hierarchical RAG & \textit{RAPTOR~\cite{sarthi2024raptor}} & \textcolor{green!60!black}{\ding{51}} & \textcolor{green!60!black}{\ding{51}} (Raw) & \textcolor{green!60!black}{\ding{51}} & \textcolor{red}{\ding{55}} ($\mathcal{O}(k \log N)$) & \textcolor{green!60!black}{\ding{51}} & \textcolor{red}{\ding{55}} (Variable) & \textcolor{red}{\ding{55}} \\
Extreme Compression & \textit{xRAG~\cite{cheng2024xragextremecontextcompression} (1-Token)} & \textcolor{green!60!black}{\ding{51}} & \textcolor{red}{\ding{55}} & \textcolor{red}{\ding{55}} & \textcolor{green!60!black}{\ding{51}} & \textcolor{green!60!black}{\ding{51}} & \textcolor{green!60!black}{\ding{51}} ($\mathcal{O}(1)$) & \textcolor{red}{\ding{55}} \\
\midrule
Latent Compressors & \textit{Gist~\cite{mu2023learning}, ICAE~\cite{ge2023icae}} & \textcolor{red}{\ding{55}} & \textcolor{green!60!black}{\ding{51}} (Latent) & \textcolor{red}{\ding{55}} & \textcolor{green!60!black}{\ding{51}} & \textcolor{green!60!black}{\ding{51}} & \textcolor{red}{\ding{55}} (Variable) & \textcolor{red}{\ding{55}} \\
Test-Time Compilation & \textit{LCC~\cite{li2026latentcontextcompilationdistilling}} & \textcolor{red}{\ding{55}} & \textcolor{green!60!black}{\ding{51}} (Buffer) & \textcolor{red}{\ding{55}} & \textcolor{green!60!black}{\ding{51}} & \textcolor{red}{\ding{55}} (Test-Time Opt.) & \textcolor{green!60!black}{\ding{51}} ($\mathcal{O}(1)$) & \textcolor{red}{\ding{55}} \\
SSM State Souping & \textit{PICASO~\cite{liu2025picaso}, Soupability~\cite{jafari2026soupability}} & \textcolor{red}{\ding{55}} & \textcolor{green!60!black}{\ding{51}} (State) & \textcolor{red}{\ding{55}} (Intra-model) & \textcolor{green!60!black}{\ding{51}} & \textcolor{green!60!black}{\ding{51}} & \textcolor{green!60!black}{\ding{51}} ($\mathcal{O}(1)$) & \textcolor{red}{\ding{55}} \\
\midrule
\rowcolor{blue!5}
\textbf{A2A Protocol} & \textbf{MII (Ours)} & \textbf{\textcolor{green!60!black}{\ding{51}}} & \textbf{\textcolor{green!60!black}{\ding{51}}} & \textbf{\textcolor{green!60!black}{\ding{51}}} & \textbf{\textcolor{green!60!black}{\ding{51}}} & \textbf{\textcolor{green!60!black}{\ding{51}} (Forward Pass)} & \textbf{\textcolor{green!60!black}{\ding{51}} ($\mathcal{O}(1)$ Fixed)} & \textbf{\textcolor{green!60!black}{\ding{51}}} \\
\bottomrule
\end{tabular}%
}
\end{table*}

%% file: sections/6_discussion.tex
\section{Conclusion and Limitations}
\label{sec:conclusion_limitations}

Modern AI is constrained by a human-centric bottleneck: relying on raw, low-density text as the universal machine memory interface. We thus ask a fundamental question: \emph{What is the native file format for neural networks?} To answer this, we introduce \textbf{Machine-Interpretable Information (MII)}, the first agent-to-agent (A2A) document-to-state protocol. By decoupling writing from reading, MII elevates documents from private hidden activations into explicit, cross-model artifacts, natively unifying searchable routing and generative memory payloads within a fixed bandwidth.

\textbf{Limitations.} While outlining a path toward fully latent machine reading, MII has current limitations:
\begin{itemize}[leftmargin=1.2em, itemsep=0pt, topsep=2pt]
    \item \textbf{Empirical Gap to Pure Losslessness:} Information-theoretically, 56 \texttt{bfloat16} tokens provide $\sim 3.67$ Mbit of bandwidth, yielding the Shannon capacity to encapsulate over 1 million text tokens. Despite this theoretical capacity for losslessness, pure \texttt{.mii} currently struggles to verbatim-reconstruct rare entities without an L2 sparse text cache (\texttt{Residual-MII}). We attribute this to the limited capacity of our 0.2B proof-of-concept Writer and alignment friction in shallow Translators.
    \item \textbf{Rigid Bandwidth Bottleneck:} Enforcing a strict $K=56$ bottleneck benefits batched predictability but forces over-compression of dense documents and under-utilization of sparse ones, lacking Variable Bitrate (VBR) flexibility.
    \item \textbf{Parametric Memory Entanglement:} It remains challenging to fully isolate knowledge decoded from the \texttt{.mii} state from the Reader's vast parametric memory~\cite{roberts2020how}. While latent transplantation strongly indicates genuine state-driven decoding, future evaluations on counterfactual corpora~\cite{longpre2021entity} are necessary to definitively disentangle protocol reconstruction from parametric hallucination.
\end{itemize}

\textbf{Broader Implications.} As scaling laws tighten the semantic sufficiency bound, residual text will diminish. MII unlocks transformative directions (Appendix~\ref{app:extended_vision}) like \textbf{Corpus-Scale Context} (a 100M-token window~\cite{chen2026msamemorysparseattention} of \texttt{.mii} encompasses $\sim$17.8B raw tokens) and \textbf{Encrypted A2A Distribution}.

As agents scale, raw text becomes an untenable transport. C2C~\cite{fu2026c2c} transfers caches pairwise; MII compiles once and decodes by many: the artifact, not the channel, is the unit of exchange. \emph{Refusing to forget is refusing to understand}. By trading lexical exactness for semantic portability, MII shifts toward \textbf{Internalization}---reading once, digesting concepts, and reasoning from compiled latent memory. We hope this is a step toward universally readable neural artifacts.

%% file: sections/8_appen.tex
\appendix

\section{Implementation and Training Details}
\label{app:implementation}

To ensure full reproducibility and to clarify the computational footprint of the MII protocol, we provide comprehensive details regarding the model architecture, training corpora, optimization hyperparameters, and evaluation configurations.

\subsection{Stage 1: Spectral Universal Writer}
\label{app:stage1_details}

\paragraph{Architecture Configurations.} 
The Writer is built upon a dual-branch Mamba architecture (Slow and Fast branches) to natively process long contexts in $\mathcal{O}(N)$ time. To enforce strict vocabulary decoupling from the modern Readers (e.g., Llama-3, Qwen), the Writer processes documents using the legacy GPT-2 tokenizer (vocab size $\approx$ 50K). Each Mamba branch consists of 12 layers with a hidden dimension of $d_{model} = 1024$, state expansion factor of 2, and $d_{state} = 16$. 

To explicitly induce the spectral specialization without unstable architectural modifications, we initialize the step size projection biases ($\Delta$) differently for each branch: $-1.5$ for the slow branch and $+0.5$ for the fast branch. The outputs of the two branches are concatenated into a fusion dimension of 2048 before routing into the 56-token memory banks (16 Global, 24 Event, 16 Cover). The Chunk Decoder utilized exclusively during Stage-1 training is a lightweight 3-layer, 8-head standard Transformer decoder.

\paragraph{Progressive Training and Regularization.}
The Writer is trained on a generalized long-document corpus comprising QA contexts and encyclopedic articles. To stabilize the training of the recurrent state space across extremely long sequences, we employ a \emph{Progressive Sequence Length Scaling} strategy. The maximum document length naturally progresses through 6 stages up to a maximum of 32,168 tokens. The models evaluated in our main experiments are trained up to Stage 3 (max 12,294 tokens).

The model is optimized using AdamW with a peak learning rate of $2 \times 10^{-4}$ (following 200 linear warmup steps) and a weight decay of $0.01$. The effective batch size is 16 (achieved via gradient accumulation). We apply gradient clipping at a maximum norm of 1.0. The bespoke spectral and stability regularizations described in Section 2 are strictly enforced: Smooth Upper Bound ($\alpha=0.6, \lambda=0.01$), Slow Floor ($\delta=0.5, \lambda=0.02$), Separation Margin ($m=0.5, \lambda=0.01$), and Dropout-based Stability Penalty ($\lambda=0.02$). Crucially, these spectral regularizations act purely as initial scaffolding: within the first 500 warmup steps, the penalty losses converge near zero. The model cleanly bifurcates and spontaneously sustains the dual-timescale routing topology without imposing long-term friction on the primary semantic objectives.

\subsection{Stage 2: Protocol Translator and Parameter Efficiency}
\label{app:stage2_details}

A core design principle of MII is shifting the heavy computational burden to the offline Writer. Crucially, the 3-layer reconstruction decoder used during Stage-1 training is entirely discarded during Stage-2 and downstream inference. Thus, the active inference components of the \textbf{Writer} (comprising the embeddings, the dual 12-layer Mamba branches, and the global attention pooling) contain only $\sim$\textbf{226M} parameters.

The Stage-2 \textbf{Translator} is designed to project the 2048-dimensional canonical MII state into the respective frozen Reader's embedding space (e.g., 4096-d for Llama-3-8B). For our primary experiments, the Translator consists of an initial linear projection followed by 2 standard Transformer encoder layers. Due to the Reader's massive hidden dimension ($d=4096$), these shallow layers (particularly the Feed-Forward networks mapping $4096 \rightarrow 8192 \rightarrow 4096$) consume $\sim$\textbf{277M} parameters.

In total, the MII continuous protocol introduces roughly \textbf{0.5B} trainable parameters. This is an order of magnitude smaller than autoencoder-based baselines like ICAE~\cite{ge2023icae}, which typically dedicate a full 7B-parameter LLM exclusively to context encoding. Furthermore, the forward pass of our 0.5B compilation pipeline takes mere milliseconds, strictly preserving the $\mathcal{O}(K)$ amortized efficiency advantage.

\section{Evaluation Configurations}
\label{app:eval_details}

\paragraph{Dataset and Context Formatting.}
For the main results on HotpotQA (Table 1), we evaluate on the full official development set comprising 7,405 queries to ensure statistical robustness. The text contexts for the \texttt{FullContext} and \texttt{Oracle} baselines, as well as the sparse evidence for \texttt{Residual-MII}, are explicitly formatted using clear role demarcations (e.g., "Context: [text] \textbackslash{}n Question: [query]") to prevent prompt injection. For \texttt{BM25} and \texttt{Dense} retrieval, documents are chunked into 90-word segments with an overlap of 25 words, and the top-3 chunks are retrieved to form a $\sim$480-token residual context.

\paragraph{LLM Generation Settings.}
To ensure strict and fair comparisons across all methods, the target LLM Readers (e.g., Meta-Llama-3-8B-Instruct) are deployed with identical generation hyperparameters. We use greedy decoding (temperature = 0.0, top\_p = 1.0) with a maximum generation length of 32 tokens, utilizing the official chat templates provided by the model authors. The generated strings are rigorously cleaned by removing padding tokens, system prompt artifacts, and redundant conversational prefixes (e.g., "Based on the text, the answer is...") before applying the official Exact Match (EM) and F1 normalization functions.

\section{Extended Vision: Theoretical Capacity and Security}
\label{app:extended_vision}

While the main text focuses on the empirical validation of the \emph{doc2state} concept under current architectural constraints, the underlying MII paradigm fundamentally alters the capacity mathematics of Large Language Models. Here, we outline the theoretical limits and broader commercial implications of compiled neural protocols.

\subsection{Information-Theoretic Capacity: Reaching 17.8 Billion Tokens}

The current industry trajectory focuses on drastically expanding LLM context windows, with recent breakthroughs such as Memory Sparse Attention (MSA) \citep{chen2026msamemorysparseattention} scaling the active limit to 100 million tokens. However, filling such massive windows with raw natural language remains fundamentally inefficient due to the inherent low information density of human-to-human communication.

By shifting the bottleneck to an offline compilation phase, MII redefines memory density. Information-theoretically, 56 continuous tokens at \texttt{bfloat16} precision provide a bandwidth of:
$$ 56 \text{ tokens} \times 4096 \text{ dimensions} \times 16 \text{ bits} \approx 3.67 \text{ Megabits (Mbit)} $$

Assuming a conservative text entropy of $\sim 2.5$ to $3$ bits per discrete token (derived from standard LLM perplexity bounds), this fixed continuous manifold possesses the theoretical Shannon capacity to encode over \textbf{1 million discrete text tokens}. This represents a potential sequence compression ratio approaching $1:20,000$.

If we extrapolate this extreme density to an architecture like MSA \citep{chen2026msamemorysparseattention}, the implications are profound. A 100M-token active window, rather than holding 100 million raw words, could instead be populated by approximately 1.78 million concatenated 56-token \texttt{.mii} states. If each state robustly encodes a 10K-token document, a single inference window would simultaneously encompass roughly \textbf{17.8 billion raw tokens}. 

This volume is equivalent to the entirety of the English Wikipedia. Thus, rather than treating "infinite context" as a hardware brute-force challenge, the MII paradigm demonstrates that true corpus-scale, multi-document reasoning can be achieved by decoupling physical context limits from cognitive capacity.
\section{The \texttt{.mii} File Format Specification (v1.0 Draft)}
\label{app:mii_schema}

To concretize the claim of MII as an explicit, transferable Agent-to-Agent (A2A) protocol, we define the preliminary \texttt{.mii} file format schema (v1.0). Unlike traditional raw text files (\texttt{.txt}) or monolithic model weights (\texttt{.pt}/\texttt{.safetensors}), the \texttt{.mii} format is designed as a lightweight, standalone artifact representing the cognitive state of a single document.

A standard \texttt{.mii} file consists of a human-readable JSON header for metadata and routing, followed by a compiled binary payload for the continuous latent memory.

\subsection{File Structure Overview}

\begin{lstlisting}[basicstyle=\ttfamily\small, frame=single, breaklines=true, numbers=none]
{
  "magic": "MII\x01",
  "version": "1.0",
  "metadata": {
    "document_id": "doc_948a7b",
    "timestamp": "2026-05-08T02:00:00Z",
    "writer_architecture": "MIIWriterv0.1(Spectral-Mamba-0.2B)",
    "writer_vocab_base": "GPT-2-Legacy"
  },
  "topology_spec": {
    "K_tokens": 56,
    "hidden_dim": 2048,
    "dtype": "bfloat16",
    "bank_allocation": {"global": 16, "event": 24, "cover": 16}
  },
  "routing_index": {
    // d=2048 Mean-pooled, for dense O(1) search
    "centroid_vector": [0.12, -0.45, ...], 
    "l2_norm": 1.0
  },
  "security_and_integrity": {
    "checksum_sha256": "8f4e2d...",
    // Reserved for encrypted A2A distribution
    "crypto_salt_id": null,       
    // Reserved for authenticated compilation
    "signature": null             
  },
  "residual_cache_optional": {
    "num_chunks": 3,
    "sparse_payload_utf8": ["chunk1...", "chunk2..."]
  },
  "__BINARY_PAYLOAD_OFFSET__": 2048
}
[... 56 x 2048 x 2 bytes (bfloat16) Continuous Latent Tensor ...]
\end{lstlisting}

\subsection{Design Rationale}

\begin{itemize}[leftmargin=1.2em]
    \item \textbf{Searchable Header:} The \texttt{routing\_index} stores the mean-pooled centroid of the \texttt{.mii} state. This allows database engines (e.g., FAISS, Milvus) to parse the header and perform highly efficient $\mathcal{O}(1)$ dense retrieval without ever loading the heavy 56-token continuous payload into GPU memory.
    \item \textbf{Type Safety and Dimensionality:} The \texttt{topology\_spec} ensures the Translator can assert dimensionality matches before attempting to inject the state into a Reader LLM, preventing silent tensor-shape errors during cross-model execution.
    \item \textbf{Residual Encapsulation:} If the state is compiled under the \texttt{Residual-MII} regime, the sparse local evidence is packaged directly within the \texttt{residual\_cache\_optional} block. This ensures the cognitive scaffold and its lexical anchors are transported as a single inseparable file.
    \item \textbf{Forward-Compatible Security:} The \texttt{security\_and\_integrity} block includes placeholders for cryptographic salting. While not empirically stress-tested in this foundational work, this schema explicitly anticipates a future where \texttt{.mii} files are encrypted and digitally signed, enabling proprietary knowledge to be safely distributed to edge Agents.
\end{itemize}

\subsection{Agent-to-Agent Protocols and Encrypted Digital Assets}

As AI agents increasingly collaborate autonomously, transmitting raw text via traditional APIs is not only computationally wasteful but also poses severe data privacy and intellectual property risks. 

The \texttt{.mii} state is continuous, machine-interpretable, and geometrically opaque to unauthenticated models or human interception. Future work in the A2A protocol space may explore \emph{cryptographic conditioning} during the Writer's compilation phase. For example, a unique cryptographic "salt" or enterprise-specific continuous key could be injected into the recurrent dynamics of the Writer. The resulting compiled artifact would be fundamentally undecodable unless the receiving Translator is equipped with the corresponding inverse alignment weights.

This creates "Compiled Knowledge Packages". It would allow enterprise knowledge bases, sensitive legal corpora, or proprietary codebases to be compiled and securely distributed over public networks. These states could be directly executed by authenticated Translator nodes at the edge, introducing a novel paradigm for encrypted digital asset distribution and proprietary RAG deployment in the AI era.

\section{Theoretical Foundations: Spectral Decoupling in State Space Models}
\label{app:math_proofs}

In Section~\ref{sec:spectral_theory}, we introduced the proposition that enforcing structural priors on the discretization step size $\Delta$ naturally leads to spectral separation, thereby minimizing gradient interference between the slow (Global) and fast (Event) channels. Here, we provide the formal continuous-time derivation and the Parseval-based proof sketch.

\subsection{Continuous-Time Frequency Response}
A linear State Space Model (SSM) operates on a continuous-time latent state $h(t) \in \mathbb{R}^N$, governed by the ODE:
\begin{align}
    \dot{h}(t) &= \mathbf{A} h(t) + \mathbf{B} x(t) \\
    y(t) &= \mathbf{C} h(t)
\end{align}
Taking the Laplace transform (with zero initial conditions), the transfer function $G(s)$ mapping the input $X(s)$ to the state $H(s)$ is:
\begin{equation}
    H(s) = (s\mathbf{I} - \mathbf{A})^{-1} \mathbf{B} X(s)
\end{equation}
Evaluating this on the imaginary axis ($s = j\omega$) gives the frequency response of the state channels. 

\subsection{The Effect of $\Delta$ Initialization on Pole Shifting}
In the discretized Mamba formulation, the continuous matrices are transformed using a step size $\Delta$. Using the zero-order hold (ZOH) approximation:
\begin{equation}
    \mathbf{\bar{A}} = \exp(\Delta \mathbf{A}), \quad \mathbf{\bar{B}} = (\Delta \mathbf{A})^{-1}(\exp(\Delta \mathbf{A}) - \mathbf{I}) \cdot \Delta \mathbf{B}
\end{equation}
Crucially, the magnitude of $\Delta$ inversely scales the effective cutoff frequency of the system. 
\begin{itemize}
    \item For the \textbf{Slow Branch} ($\Delta_{\mathrm{slow}}$ initialized very small via $-1.5$ bias), the system acts as a strict \emph{low-pass filter}. It attenuates high-frequency local fluctuations and integrates long-range semantic shifts.
    \item For the \textbf{Fast Branch} ($\Delta_{\mathrm{fast}}$ initialized large via $+0.5$ bias), the system acts more like a \emph{high-pass/band-pass filter}, remaining highly responsive to immediate, token-to-token lexical transitions (events).
\end{itemize}

\subsection{Gradient Decoupling via Parseval's Theorem}
During backpropagation through time (BPTT), the gradient signals $\nabla_{\theta_{\mathrm{slow}}} \mathcal{L}$ and $\nabla_{\theta_{\mathrm{fast}}} \mathcal{L}$ propagate through the respective state trajectories $h_{\mathrm{slow}}(t)$ and $h_{\mathrm{fast}}(t)$.

To measure the interference (cross-talk) between these learning signals, we examine their inner product in the time domain. According to \textbf{Parseval's Theorem}, the integral of the product of two time-domain signals is proportional to the integral of the product of their Fourier transforms:
\begin{equation}
    \int_{-\infty}^{\infty} h_{\mathrm{slow}}(t) \cdot h_{\mathrm{fast}}(t) \, dt = \frac{1}{2\pi} \int_{-\infty}^{\infty} \hat{H}_{\mathrm{slow}}(\omega) \cdot \hat{H}_{\mathrm{fast}}^*(\omega) \, d\omega
\end{equation}
Because our $\Delta$ initialization explicitly shifts the spectral support (frequency bands) of $\hat{H}_{\mathrm{slow}}(\omega)$ and $\hat{H}_{\mathrm{fast}}(\omega)$ to be nearly disjoint (as empirically verified in Figure~\ref{fig:dtbias}), their cross-spectral density $\hat{H}_{\mathrm{slow}}(\omega) \cdot \hat{H}_{\mathrm{fast}}^*(\omega)$ approaches zero across almost all frequencies $\omega$.

Therefore, the time-domain inner product of the states—and correspondingly, the expected inner product of their backpropagated gradients—is minimized:
\begin{equation}
    \mathbb{E}\left[ \langle \nabla_{\theta_{\mathrm{slow}}} \mathcal{L}, \nabla_{\theta_{\mathrm{fast}}} \mathcal{L} \rangle \right] \approx 0
\end{equation}
This theoretical guarantee ensures that the MII Writer can simultaneously compress macroscopic summaries (Global bank) and microscopic details (Event bank) into the same 56-token manifold without catastrophic interference.

\section{Extended Ablation: The Scaling Law of Latent Compression}
\label{app:scaling_laws}

In the main text (Section~\ref{sec:conclusion_limitations}), we argued that the current empirical gap to pure semantic losslessness is primarily a function of model capacity, rather than a theoretical flaw in the protocol. Here, we provide the empirical scaling curves to support this claim.

\begin{figure}[htbp]
    \centering
    \includegraphics[width=0.6\linewidth]{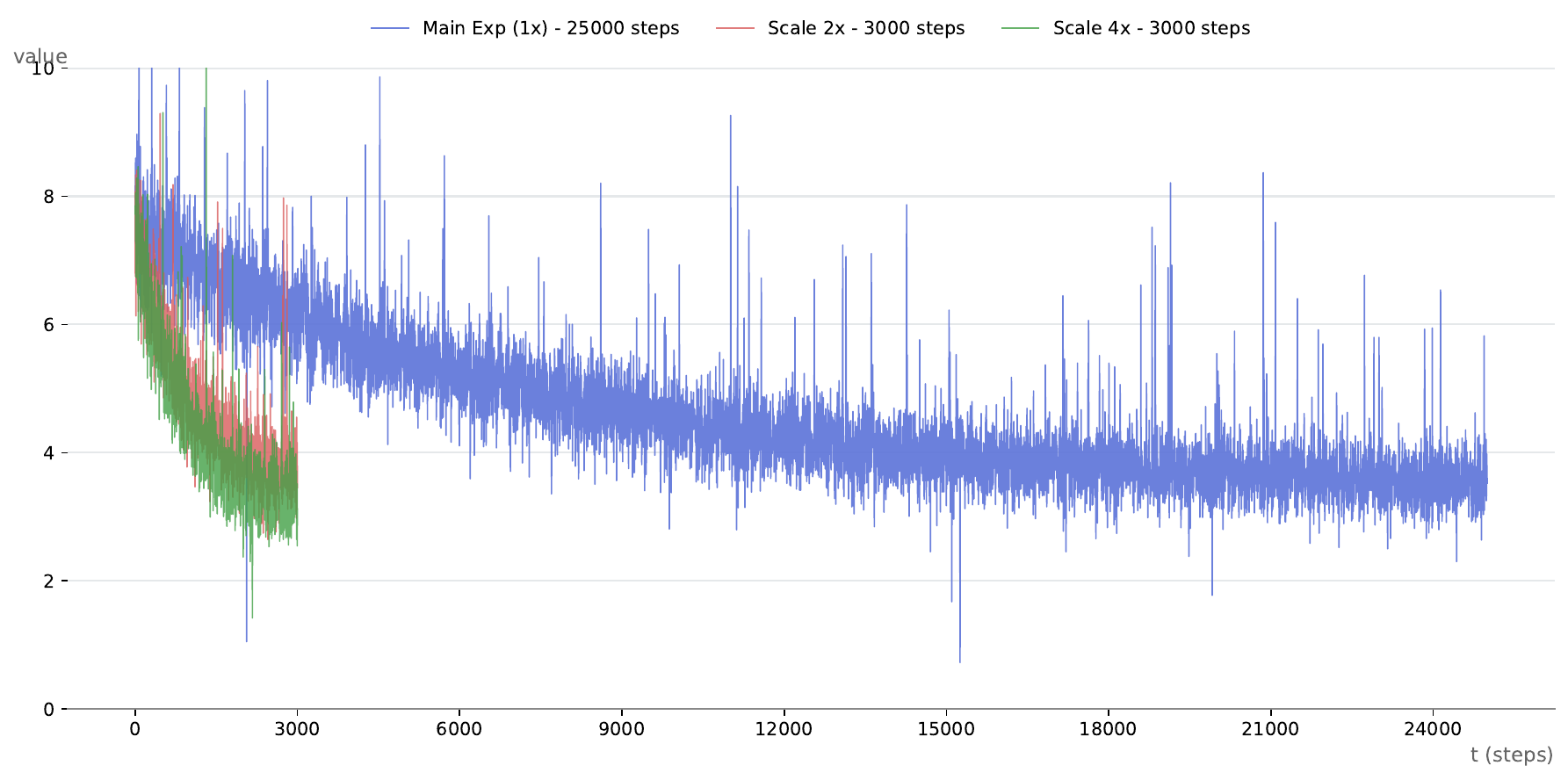}
    \caption{\textbf{Writer Scaling Dynamics.} As the Writer capacity scales from 1x to 4x, the reconstruction loss over the fixed 56-token protocol state decays exponentially faster, confirming that semantic compression follows predictable scaling laws.}
    \label{fig:app_scaling}
\end{figure}

As shown in Figure~\ref{fig:app_scaling}, we evaluated the Stage-1 Context Reconstruction Loss ($\mathcal{L}_{\mathrm{rec}}$) across different Writer capacities while maintaining the exact same 56-token protocol bandwidth. 
\begin{itemize}
    \item The baseline 1x Writer (0.2B parameters) converges slowly, reaching a reconstruction loss of $\sim 3.53$ at 25,000 steps.
    \item The 2x capacity Writer drops to a loss of 3.74 in merely 3,000 steps.
    \item The 4x capacity Writer achieves an even steeper trajectory, plummeting to a loss of 3.19 at 3,000 steps.
    \item \texttt{(The loss is calculated by the nearest 100 steps loss output)}
\end{itemize}

This power-law decay strongly implies that extracting high-frequency textual details into a fixed continuous manifold is heavily bottlenecked by the Writer's functional approximation capacity. The protocol interface ($K=56$) is not the limit; the parameter count is. We project that training an MII Writer at the 7B parameter scale will completely close the gap between \texttt{Residual-MII} and \texttt{Pure MII}, achieving true verbatim generative fidelity.

\section{Qualitative Transcripts: Emergent 5-Hop Latent Reasoning}
\label{app:qualitative_5hop}

To vividly illustrate the compositional power of the \texttt{.mii} protocol, we provide the complete transcript of the 5-hop latent reasoning test discussed in Section~\ref{sec:white_box}. In this test, the Reader is provided \textbf{zero raw text}. It is fed only the concatenation of two independently compiled MII states.

\subsection{The Hidden Logical Chain}
The test requires traversing the following 5 hops across two separate documents:
\begin{enumerate}
    \item Who invented the Aero-Lattice framing technique for the Cathedral of Glass? $\rightarrow$ \emph{Tobias Sterling} (Found in Doc 1).
    \item What is the final sculpture created by Tobias Sterling? $\rightarrow$ \emph{A kinetic sculpture of obsidian and silver} (Found in Doc 2).
    \item Who was it entrusted to? $\rightarrow$ \emph{His estranged niece, Eleanor} (Found in Doc 2).
    \item Who did Eleanor marry? $\rightarrow$ \emph{A Japanese diplomat} (Found in Doc 2).
    \item Where did she place the sculpture? $\rightarrow$ \textbf{\emph{The Sapphire Pavilion in Kyoto}} (Found in Doc 2).
\end{enumerate}

\subsection{Model Outputs (Zero-Shot)}

\textbf{Prompt Format given to the frozen Llama-3-8B-Instruct:}
\begin{verbatim}
<|begin_of_text|><|start_header_id|>system<|end_header_id|>
You are a helpful AI assistant.<|eot_id|>
<|start_header_id|>user<|end_header_id|>
[MII_STATE_DOC1][MII_STATE_DOC2]
According to the provided memories, what is the current location of the 
final sculpture created by the individual who invented the Aero-Lattice 
framing technique used in the Cathedral of Glass?<|eot_id|>
<|start_header_id|>assistant<|end_header_id|>
\end{verbatim}

\textbf{Result 1: Blind (No MII States provided)}
\begin{quote}
\textit{"The current location of the final sculpture created by the individual who invented the Aero-Lattice framing technique used in the Cathedral of Glass is not specified in standard historical records, as this appears to be a fictional premise..."} (Hallucination / Safety Abstention)
\end{quote}

\textbf{Result 2: Single State (Only [MII\_STATE\_DOC1] provided)}
\begin{quote}
\textit{"Based on the memory, the inventor is Tobias Sterling. However, the text only mentions his work on the Cathedral of Glass in London. It does not contain information about his final sculpture or its location."} (Correctly identifies missing information)
\end{quote}

\textbf{Result 3: Concatenated States ([MII\_STATE\_DOC1] + [MII\_STATE\_DOC2])}
\begin{quote}
\textit{"According to the integrated memories, Tobias Sterling invented the Aero-Lattice technique. His final creation, an obsidian and silver kinetic sculpture, was transported by his niece Eleanor to her new home. It is currently permanently housed in the private archives of the \textbf{Sapphire Pavilion in Kyoto}."} (\textbf{Success: Accurate 5-hop latent deduction})
\end{quote}

This result confirms that \texttt{.mii} states are not merely lossy text summaries, but highly structured cognitive manifolds that permit the Reader's attention heads to dynamically route logical queries across independently compiled semantic topologies.

\section{Extended Related Work}
\label{app:extended_related_work}

Due to space constraints in the main text, Section~\ref{sec:related_work} focused specifically on methodologies directly adjacent to context compression and protocol paradigms. Here, we provide a comprehensive review of the broader literature contextualizing the development of MII.

\subsection{Evolution of Retrieval-Augmented Generation (RAG)}
The paradigm of decoupling knowledge storage from parametric memory was popularized by standard dense retrieval systems like DPR~\cite{karpukhin2020dense} and REALM~\cite{guu2020realm}. Subsequent architectures sought to better integrate retrieved passages with generative models, notably FiD (Fusion-in-Decoder)~\cite{izacard2020leveraging} and Atlas~\cite{izacard2022atlas}, which process multiple retrieved chunks in parallel. More granular retrieval techniques, such as ColBERT's late interaction~\cite{khattab2020colbert}, HyDE's pseudo-document generation~\cite{gao2022precise}, and Self-RAG's critique mechanisms~\cite{asai2023self}, have significantly improved retrieval precision. However, all these methods inherit the fundamental index--payload schism: they rely on vectors for search but require the LLM to process raw string payloads for reasoning. MII fundamentally diverges by compiling documents into a protocol state that natively functions as both the dense index and the generative memory.

\subsection{Long-Context and Compressive Sequence Modeling}
To mitigate the $\mathcal{O}(N^2)$ attention bottleneck of Transformers, substantial work has explored sparse attention mechanisms (e.g., Longformer~\cite{beltagy2020longformer}) and advanced position encodings (e.g., ALiBi~\cite{press2021train}, RoPE~\cite{su2024roformer}). Despite these advances allowing windows of up to 100M tokens, LLMs still suffer from the "Lost in the Middle" phenomenon~\cite{liu2024lost}, a fragility increasingly exposed by extreme stress-test benchmarks like \textbf{LEval}~\cite{an2023leval} and \textbf{Needle In A Haystack (NIAH)}~\cite{kamradt2023needle}. 
Compressive memory models, such as Compressive Transformers~\cite{rae2019compressive}, Memorizing Transformers~\cite{wu2022memorizing}, RMT~\cite{bulatov2022recurrent}, and LongLLaMA~\cite{tworkowski2023focused}, attempt to store past contexts in compact continuous vectors or differentiable kNN memory banks. MII shares the philosophical goal of continuous memory, but shifts the computational paradigm: rather than computing memory online during inference, MII utilizes a dedicated, highly efficient Writer to compile the memory offline, yielding a portable state.

\subsection{KV-Cache Management and Lexical Pruning}
To handle the massive memory footprint of long contexts, orthogonal system-level approaches attempt to sparsify or quantize the KV-cache of a \emph{single} model during inference. Methods like \textbf{H2O}~\cite{zhang2023h2o} and \textbf{SnapKV}~\cite{li2024snapkv} aggressively evict non-essential tokens based on historical attention scores, whereas \textbf{StreamingLLM}~\cite{xiao2023efficient} preserves initial attention sinks for infinite generation. Alternatively, \textbf{KIVI}~\cite{liu2024kivi} and \textbf{KVQuant}~\cite{hooper2024kvquant} compress the continuous cache precision down to 2-4 bits. On the lexical front, pruning frameworks like \textbf{Selective Context}~\cite{li2023evaluating} shorten raw prompts by filtering tokens via information entropy. While these methods excellently optimize single-model inference, they still require re-processing raw text globally and are inextricably bound to specific architectures, contrasting with MII's architecture-agnostic A2A protocol design.

\subsection{Parameter-Efficient Tuning and Cross-Space Alignment}
MII's use of a lightweight Translator to map canonical states into specific LLM embedding spaces intersects with prompt tuning and representation alignment. Continuous prompt learning methods like Prefix-Tuning~\cite{li2021prefix} and Prompt Tuning~\cite{lester2021power} demonstrated that frozen LLMs can be steered by prepended continuous vectors. Our Translator optimization is related, but differs fundamentally in objective: rather than learning a single static task prefix, the Translator learns a dynamic mapping function to bridge diverse document topologies. 
The feasibility of this cross-model mapping is grounded in literature on neural representation similarity, such as SVCCA~\cite{raghu2017svcca} and CKA~\cite{kornblith2019similarity}. Recent works on relative representations~\cite{moschella2022relative} demonstrate that distinct neural networks often learn isometric latent spaces. MII exploits this phenomenon, proving that a universally compiled document manifold can be efficiently stitched into diverse target spaces.

\subsection{Mechanistic Interpretability and Knowledge Editing}
Our latent space white-boxing experiments (Section~\ref{sec:white_box}) are heavily inspired by recent advancements in mechanistic interpretability. Tools for causal tracing, pioneered by ROME~\cite{meng2022locating} and MEMIT~\cite{meng2022mass}, enable the precise localization of factual associations within Transformer weights. Furthermore, the understanding that feed-forward networks act as key-value memories~\cite{geva2020transformer} provides a theoretical basis for why our \texttt{.mii} Event and Cover banks successfully induce specific entity generation. MII extends these probing techniques to continuous prompt states, demonstrating that compressed artifacts maintain structured, editable semantic topologies rather than entangled black-box representations.